\documentclass[11pt,a4paper]{article}
\usepackage[hyperref]{acl2019}
\usepackage{times}
\usepackage{latexsym}
\usepackage{booktabs}
\usepackage{amsmath}
\usepackage[utf8]{inputenc}
\usepackage{url}

\usepackage{multirow}
\usepackage{rotating}
\usepackage[frozencache,cachedir=.]{minted}

\aclfinalcopy 

\newenvironment{itemizesquish}{\begin{list}{\labelitemi}{\setlength{\topsep}{0.5em}\setlength{\itemsep}{0em}\setlength{\labelwidth}{0.5em}\setlength{\leftmargin}{\labelwidth}\addtolength{\leftmargin}{\labelsep}}}{\end{list}}

\title{OpenKiwi: An Open Source Framework for Quality Estimation}

\author{F\'abio Kepler \\
  Unbabel 
 \And
  Jonay Tr\'enous \\
  Unbabel 
  \And
  Marcos Treviso\thanks{\,\,\, Work done during an internship at Unbabel in 2018.}\\
  Instituto de Telecomunica\c{c}\~oes
 \AND 
  Miguel Vera \\
  Unbabel
 \And
  Andr\'e F.~T.~Martins \\
  Unbabel 
\AND \\ 
  \texttt{\{kepler, sony, miguel.vera, andre.martins\}@unbabel.com}\\
  \texttt{marcosvtreviso@gmail.com}
}

\date{}

\begin{document}
\maketitle
\begin{abstract}
  We introduce OpenKiwi, a PyTorch-based open source framework for translation quality estimation. OpenKiwi supports training and testing of word-level and sentence-level quality estimation systems, implementing the winning systems of the WMT 2015--18 quality estimation campaigns. We benchmark OpenKiwi on two datasets from WMT 2018 (English-German SMT and NMT), yielding state-of-the-art performance on the word-level tasks and near state-of-the-art in the sentence-level tasks.
\end{abstract}

\section{Introduction}


\textbf{Quality estimation} (QE) provides the missing link between machine and human translation: its goal is to evaluate a translation system's quality without access to reference translations~\cite{Specia2018}. 
Among its potential usages are:
informing an end user about the reliability of automatically translated content;
deciding if a translation is ready for publishing or if it requires human post-editing;
and highlighting the words that need to be post-edited.

While there has been tremendous progress in QE in the last years \cite{Martins2016,Martins2017TACL,Kim2017,Wang2018}, the ability of researchers to reproduce state-of-the-art systems has been hampered by the fact that these are either based on complex ensemble systems, complicated architectures, or require not well-documented pre-training and fine-tuning of some components.
Existing open-source frameworks such as WCE-LIG \citep{Servan2015}, QuEST++ \citep{Specia2015}, Marmot \citep{Logacheva2016}, or DeepQuest \citep{Ive2018}, while helpful, are currently behind the recent best systems in WMT QE shared tasks.
%
To address the shortcoming above, this paper presents {\bf OpenKiwi},%
\footnote{\url{https://unbabel.github.io/OpenKiwi}.} %
a new open source framework for QE that implements the best QE systems from WMT 2015--18 shared tasks, making it easy to combine and modify their key components, while experimenting under the same framework.

The main features of OpenKiwi are:
\begin{itemizesquish}
    \item Implementation of four QE systems:
    {\sc QUETCH} \citep{Kreutzer2015},
    {\sc NuQE} \citep{Martins2016,Martins2017TACL},
    Predictor-Estimator \citep{Kim2017,Wang2018},
    and a stacked ensemble with a linear system \citep{Martins2016,Martins2017TACL};
    \item Easy to use API: can be imported as a package in other projects or run from the command line;
    \item Implementation in Python using PyTorch as the deep learning framework;
    \item Ability to train new QE models on new data;
    \item Ability to run pre-trained QE models on data from the WMT 2018 campaign;
    \item Easy to track and reproduce experiments via {\tt YAML} configuration files and (optionally) MLflow;
    \item Open-source license (Affero GPL).
\end{itemizesquish}

This project is hosted at \url{https://github.com/Unbabel/OpenKiwi}.
We welcome and encourage contributions from the research community.%
\footnote{See \url{https://unbabel.github.io/OpenKiwi/contributing.html} for instructions for contributors.}

\begin{figure*}[!t]
\small
\centering
\includegraphics[width=\textwidth]{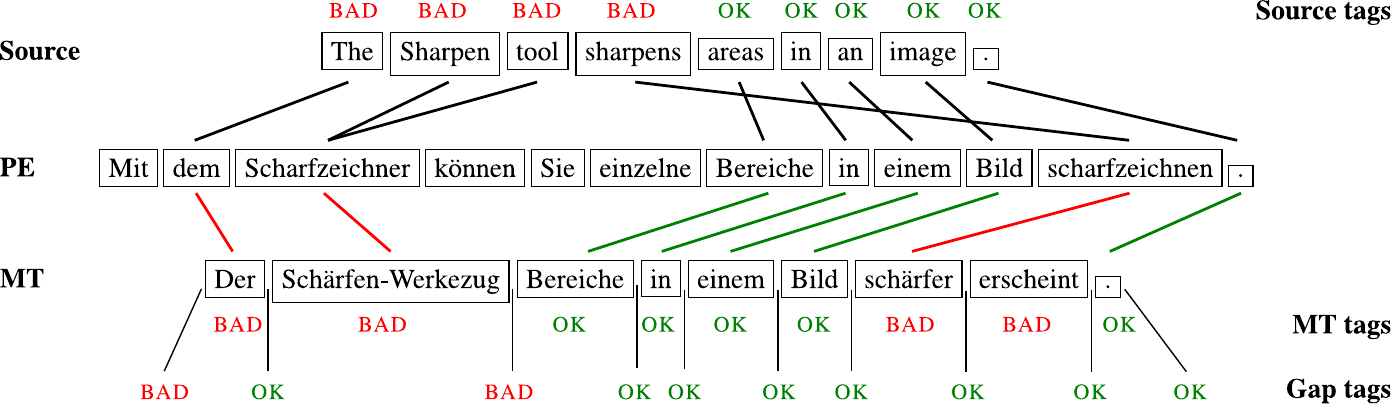}
\caption{\label{fig:qe}Example from the WMT 2018 word-level QE training set. Shown are the English source sentence (top), the German machine translated text (bottom), and its manual post-edition (middle). We show also the three types of word-level quality tags: MT (or target) tags account for words that are replaced or deleted, gap tags account for words that need to be inserted, and source tags indicate what are the source words that were omitted or mistranslated. For this example, the HTER sentence-level score (number of edit operations to produce PE from MT normalized by the length of PE) is $8/12 = 66.7\%$, corresponding to 4 insertions, 1 deletion, and 3 replacements out of 12 reference words.}
\end{figure*}

\section{Quality Estimation}

The goal of {\bf word-level QE} (Figure~\ref{fig:qe}) is to assign quality labels (\textsc{ok} or \textsc{bad}) to each \emph{machine-translated word}, as well as to \emph{gaps} between words (to account for context that needs to be inserted), and \emph{source words} (to denote words in the original sentence that have been mistranslated or omitted in the target). 
In the last years, the most accurate systems that have been developed for this task combine linear and neural models
\cite{Kreutzer2015,Martins2016}, 
use automatic post-editing as an intermediate step \citep{Martins2017TACL}, or develop specialized neural architectures \citep{Kim2017,Wang2018}. 

{\bf Sentence-level QE}, on the other hand, aims to predict the quality of the whole translated sentence, for example based on the time it takes for a human to post-edit it, or on how many edit operations are required to fix it, in terms of HTER (Human Translation Error Rate) \citep{Specia2018}. 
The most successful approaches to sentence-level QE to date are based on conversions from word-level predictions \citep{Martins2017TACL} or joint training with multi-task learning \citep{Kim2017,Wang2018}.

\section{Implemented Systems}

OpenKiwi implements four popular systems that have been proposed in the last years, which we now describe briefly.

\paragraph{QUETCH.} The ``QUality Estimation from scraTCH'' system \citep{Kreutzer2015} is designed as a multilayer perceptron with one hidden layer, non-linear \textit{tanh} activation functions and a lookup-table layer mapping words to continuous dense vectors. 
For each position in the MT, a window of fixed size surrounding that position, as well as a windowed representation of aligned words from the source text, are concatenated as model input.%
\footnote{The alignments are provided by the shared task organizers, which are computed with {\tt fast\_align} \citep{Dyer2013}.} %
The output layer scores \textsc{ok/bad} probabilities for each word with a softmax activation. The model is trained independently to predict source tags, gap tags, and target tags. 
QUETCH is a very simple model and does not rely on any kind of external auxiliary data for training, only the shared task datasets.

\paragraph{NuQE.} OpenKiwi also implements the NeUral Quality Estimation system proposed by \citet{Martins2016}.
Its architecture consists of a lookup layer containing embeddings for target words and their source-aligned words, in the same fashion as QUETCH.
These embeddings are concatenated and fed into two consecutive sets of two feed-forward layers and a bi-directional GRU layer.
The output contains a softmax layer that produces the final {\sc ok}/{\sc bad} decisions.
Like QUETCH, training is also carried independently for source tags, gap tags, and target tags. 
NuQE is also a blackbox system, meaning it is trained with the shared task data only (i.e., no auxiliary parallel or roundtrip data).


\paragraph{Predictor-Estimator.} Our implementation follows closely the architecture proposed by \citet{Kim2017}, which consists of two modules:
\begin{itemizesquish}
    \item a \emph{predictor}, which is trained to predict each token of the target sentence given the source and the left and right context of the target sentence;
    \item an \emph{estimator}, which takes features produced by the \emph{predictor} and uses them to classify each word as {\sc ok} or {\sc bad}.
\end{itemizesquish} 
Our predictor uses a bidirectional LSTM to encode the source, and two unidirectional LSTMs processing the target in left-to-right (LSTM-L2R) and right-to-left (LSTM-R2L) order.
For each target token $t_i$, the representations of its left and right context are concatenated and used as query to an attention module before a final softmax layer.
It is trained on the large parallel corpora provided as additional data by the WMT shared task organizers.
The estimator takes as input a sequence of features: for each target token $t_i$, the final layer before the softmax (before processing $t_i$), and the concatenation of the $i$-th hidden state of LSTM-L2R and LSTM-R2L (after processing $t_i$).
In addition, we train this system with a multi-task architecture that allows us to predict sentence-level HTER scores.
Overall, this system is capable to predict sentence-level scores and all word-level labels (for MT words, gaps, and source words)---the source word labels are produced by training a predictor in the reverse direction.

\paragraph{Stacked Ensemble.} The systems above can be ensembled by using a stacked architecture with a feature-based linear system, as described by \citet{Martins2017TACL}. The features are the ones described there, including lexical and part-of-speech tags from words, their contexts, and their aligned words and contexts, as well as syntactic features and features provided by a language model (as provided by the shared task organizers).
This system is only used to produce word-level labels for MT words.

\section{Design, Implementation and Usage}

OpenKiwi is designed and implemented in a way that allows new models to be easily added and run, without requiring much concern about input data processing and output generation and evaluation.
That means the focus can be almost exclusively put in adding or changing a {\tt torch.nn.Module} based class.
If new flags or options are required, all that is needed is to add them to the CLI parsing module.

\paragraph{Design.}
As a general architecture example, the training pipeline follows these steps:
\begin{itemizesquish}
    \item Each input data, like source text and MT text, is defined as a \texttt{Field}, which holds information about how data should be tokenized, how the inner vocabulary is built, how the mapping to IDs is done, and how a list of samples is padded into a tensor;
    \item A \texttt{Dataset} holds a set of input and output fields, and builds minibatches of samples, each containing their respective input and output data;
    \item A training loop iterates over epochs and steps, calling the model with each minibatch, computing the loss, backpropagating, evaluating on the validation set, and saving snapshots as requested;
    \item By default, the best model is kept and predictions on the validation set are saved as probabilities.
\end{itemizesquish}

The flow rarely needs to be changed for the QE task, so all that is needed for quick experimentation is changing configuration parameters (check the \textbf{Usage} part below) or the model class.

\paragraph{Implementation.}
OpenKiwi supports Python 3.5 and later.
Since reproducibility is important, it uses Poetry\footnote{\url{https://poetry.eustace.io/}} for deterministic dependency management.
To decrease the risk of introducing breaking changes with new code, a set of tests are also implemented and currently provide a code coverage close to $80\%$.

OpenKiwi offers support for tracking experiments with MLflow,\footnote{\url{https://mlflow.org/}} which allows comparing different runs and searching for specific metrics and parameters.

\begin{figure*}[!t]
\small
\centering
\includegraphics[width=0.65\textwidth]{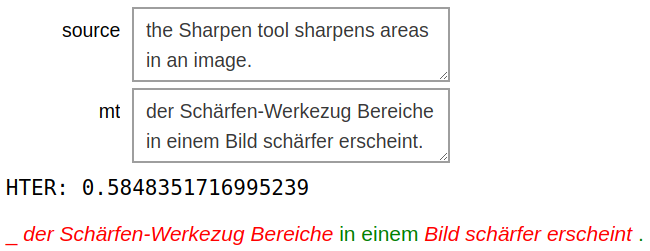}

\caption{\label{fig:ex}Interactive visualization of the system output. Words tagged as {\sc bad} as shown in \textcolor{red}{\it red}, and {\sc bad} gaps are denoted as red underscores (``\textcolor{red}{\_}"). The Jupyter Notebook producing this output is available at  \url{https://github.com/Unbabel/OpenKiwi/blob/master/demo/KiwiViz.ipynb}.}
\end{figure*}

\paragraph{Usage.}
Training an OpenKiwi model is as simple as running the following command:
\begin{minted}[breaklines]{bash}
$ python kiwi train --config config.yml
\end{minted}
where {\tt config.yml} is a configuration file with training and model options.

OpenKiwi can also be installed as a Python package by running {\tt pip install openkiwi}.
In this case, the above command can be switched by
\begin{minted}{bash}
$ kiwi train --config config.yml
\end{minted}
If used inside another Python project, OpenKiwi can be easily used like the following:
\begin{minted}{python}
import kiwi

config = 'config.yml'
run_info = kiwi.train(config)
\end{minted}

After training, predicting on new data can be performed by simply calling
\begin{minted}{python}
model = kiwi.load_model(
    run_info.model_path
)
source = [
    'the Sharpen tool sharpens '
    'areas in an image .'
]
target = [
    'der Schärfen-Werkezug '
    'Bereiche in einem Bild '
    'schärfer erscheint .'
]
examples = [{
    'source': source,
    'target': target
}]
out = model.predict(examples)
\end{minted}

Figure \ref{fig:ex} shows an example of QE predictions using the framework.


\section{Benchmark Experiments}

\begin{table*}[!t]
    \centering
\begin{tabular}{lcccccccccc}
\toprule
\multirow{2}{*}{Model} & \multicolumn{5}{c}{En-De SMT} & \multicolumn{5}{c}{En-De NMT} \\
   &        MT &  gaps & source &   $r$ & $\rho$ &        MT &  gaps & source &   $r$ & $\rho$ \\
\midrule
\textsc{QUETCH}      &     39.90 & 17.10 &  36.10 & 48.32 &  51.31 &     29.18 & 13.26 &  28.91 & 42.84 &  49.59 \\
\textsc{NuQE}        &     50.04 & 35.53 &  42.08 & 59.62 &  60.89 &     32.49 & 15.01 &  30.19 & 43.41 &  50.87 \\
\textsc{Pred-Est}    &     57.29 & 43.68 &      33.02 & 70.95 &  74.49 &     39.25 & 21.54 &      29.52 & 50.18 &  55.66 \\
\textsc{APE-QE}      &     55.12 & 47.04 &  51.11 & 58.01 &  60.58 &     37.60 & 21.78 &  {\bf 34.46} & 35.23 &  38.88 \\
\midrule
\textsc{Ensembled}   &          61.33 &     {\bf 53.05} &      {\bf 51.11} & {\bf 72.89}     & {\bf 76.37}      &         43.04 &     {\bf 24.74} &      {\bf 34.46} &     {\bf 52.34} &      {\bf 56.98} \\
\textsc{Stacked}     &         {\bf 62.40} &    -- &     -- &    -- &     -- &         {\bf 43.88} &    -- &     -- &    -- &     -- \\
\bottomrule
\end{tabular}
\caption{Benchmarking of the different models implemented in {\tt OpenKiwi} on the WMT 2018 development set, along with an ensembled system ({\sc Ensembled}) that averages the predictions of the {\sc NuQE}, {\sc APE-QE}, and {\sc Pred-Est} systems, as well as a stacked architecture ({\sc Stacked}) which stacks their predictions into a linear feature-based model, as described by \citet{Martins2017TACL}. For each system, we report the five official scores used in WMT 2018: word-level $F_1^\mathrm{mult}$ for MT, gaps, and source tokens, and sentence-level Pearson's $r$ and Spearman's $\rho$ rank correlations.}
\label{tab:results_dev}
\end{table*}

\begin{table*}[t]
    \centering
\begin{tabular}{lcccccccccc}
\toprule
\multirow{2}{*}{Model} & \multicolumn{5}{c}{En-De SMT} & \multicolumn{5}{c}{En-De NMT} \\
   &        MT &  gaps & source &   $r$ & $\rho$ &        MT &  gaps & source &   $r$ & $\rho$ \\
\midrule
{\tt deepQUEST} & 42.98 & 28.24 & \underline{33.97} & 48.72 & 50.97 & 30.31 & \underline{11.93} & \underline{28.59} & 38.08 & 48.00 \\
\texttt{UNQE} & -- & -- & -- & 70.00 & 72.44 & -- & -- & -- & \underline{\textbf{51.29}} & \underline{\textbf{60.52}} \\
\texttt{QE Brain} & \underline{62.46} & \underline{49.99} & -- & \underline{\textbf{73.97}} & \underline{\textbf{75.43}} & \underline{43.61} & -- &  -- & 50.12 & 60.49 \\
\midrule
{\tt OpenKiwi} & {\bf 62.70} & {\bf 52.14} & {\bf 48.88} & 71.08 & 72.70 & {\bf 44.77} & {\bf 22.89} & {\bf 36.53} & 46.72 & 58.51 \\
\bottomrule
\end{tabular}
\caption{\label{tab:results_test} Final results on the WMT 2018 test set.
The first three systems are the official WMT18-QE winners (underlined):
{\tt deepQUEST} is the open source system developed by \citet{Ive2018}, \texttt{UNQE} is the unpublished system from Jiangxi Normal University, described by \citet{WMT2018_findings}, and \texttt{QE Brain} is the system from Alibaba described by \citet{Wang2018}.
Reported numbers for the {\tt OpenKiwi} system correspond to best models in the development set: the {\sc Stacked} model for prediction of MT tags, and the {\sc Ensembled} model for the rest.}
\end{table*}

\paragraph{Datasets.}
To benchmark OpenKiwi, we use the following datasets from the WMT 2018 quality estimation shared task, all English-German (En-De):
\begin{itemizesquish}
    \item Two quality estimation datasets of sentence triplets, each consisting of a source sentence (SRC), its machine translation (MT) and a human post-edition (PE) of the machine translation:  
    a larger dataset of 26,273 training and 1,000 development triplets, where the MT is generated by a phrase-based statistical machine translation (SMT); 
    and a smaller dataset of 13,442 training and 1,000 development triplets, where the MT is generated by a neural machine translation system (NMT). 
    The data also contains word-level quality labels and sentence-level scores that are obtained from the post-editions using \textsc{tercom} \citep{Snover2006}. 
    \item A corpus of 526,368 artificially generated sentence triplets, obtained by first cross-entropy filtering a much larger monolingual corpus for in-domain sentences, then using round-trip translation and a final stratified sampling step.
    \item A parallel dataset of 3,396,364 in-domain sentences used for pre-training of the predictor-estimator model.
\end{itemizesquish}


\paragraph{Systems.}
In addition to the models that are part of OpenKiwi, in the experiments below, we also use Automatic Post-Editing (APE) adapted for QE (\textsc{APE-QE}).
APE-QE has been used by \citet{Martins2017TACL} as an intermediate step for quality estimation, where an APE system is trained on the human post-edits and its outputs are used as pseudo-post-editions to generate word-level quality labels and sentence-level scores in the same way that the original labels were created.
Since OpenKiwi's focus is not on implementing a sequence-to-sequence model, we used an external software, OpenNMT-py \citep{Klein2017}, to train two separate translation models: 
\begin{itemizesquish}
    \item SRC $\rightarrow$ PE: trained first on the in-domain corpus provided, then fine-tuned on the shared task data.
    \item MT $\rightarrow$ PE: trained on the concatenation of the corpus of artificially created sentence triplets and the shared task data oversampled by a factor of 20.
\end{itemizesquish} 
These predictions are then combined in the ensemble and stacked systems as explained below.



\paragraph{Experiments.}
We show benchmark numbers on the two English-German WMT 2018 datasets.
In Table~\ref{tab:results_dev}, we compare different configurations of OpenKiwi on the development datasets. For the single systems, we can see that the predictor-estimator has the best performance, except for predicting the source and the gap word-level tags, where APE-QE is superior. Overall, ensembled versions of these systems perform the best, with a stacked architecture being very effective for predicting word-level MT labels, confirming the findings of \citet{Martins2017TACL}. 

Finally, in Table~\ref{tab:results_test}, we report numbers on the official test set. 
We compare OpenKiwi against the best systems in WMT 2018 \citep{WMT2018_findings} and another existing open-source tool, {\tt deepQuest} \citep{Ive2018}. Overall, OpenKiwi outperforms {\tt deepQuest} for all word-level and sentence-level tasks, and attains the best results for all the word-level tasks.

\section{Conclusions}

We presented OpenKiwi, a new open source framework for QE. OpenKiwi is implemented in PyTorch and supports training of word-level and sentence-level QE systems on new data. It outperforms other open source toolkits on both word-level and sentence-level, and yields new state-of-the-art word-level QE results.

Since its release, OpenKiwi was adopted as the baseline system for the WMT 2019 QE shared task%
\footnote{More specifically, the NuQE model: \url{http://www.statmt.org/wmt19/qe-task.html}},
Moreover, all the winning systems of the word-, sentence- and document-level tasks of the WMT 2019 QE shared task%
\footnote{\url{http://www.statmt.org/wmt19/qe-results.html}}
\citep{Kepler2019}
used OpenKiwi as their building foundation.

\section*{Acknowledgments}

The authors would like to thank Eduardo Fierro, Thomas Reynaud, and the Unbabel AI and Engineering teams for their invaluable contributions to OpenKiwi.

We  would  also  like  to  thank  the  support  provided by the EU in the context of the PT2020 project (contracts 027767 and 038510), by  the  European  Research  Council  (ERC  StG  DeepSPIN  758969), and  by  the  Fundação  para  a  Ciência  e  Tecnologia through contract UID/EEA/50008/2019.

\bibliography{acl2019}
\bibliographystyle{acl_natbib}

\end{document}